\documentclass[10pt,twocolumn,letterpaper]{article}

\usepackage{cvpr}
\usepackage{times}
\usepackage{epsfig}
\usepackage{graphicx}
\usepackage{amsmath}
\usepackage{amssymb}
\usepackage{multirow}

\usepackage[nocompress]{cite}
\usepackage[usenames, dvipsnames]{color}

\newcommand{\name}{Deep Anytime Re-ID}
\newcommand{\nameshort}{DaRe}
\usepackage[pagebackref=true,breaklinks=true,letterpaper=true,colorlinks,bookmarks=false]{hyperref}
\renewcommand{\paragraph}[1]{\vspace{0.25ex}\noindent\textbf{#1}}
\cvprfinalcopy % *** Uncomment this line for the final submission

\def\cvprPaperID{2509} % *** 	Enter the CVPR Paper ID here

\ifcvprfinal\fi
\begin{document}

%%%%%%%%% TITLE
\title{Resource Aware Person Re-identification across Multiple Resolutions}

\author{Yan Wang\thanks{Authors contributed equally.}~~\thanks{Cornell University. \{yw763, lw633, zc346, sl2327, gh349\}@cornell.edu, bharathh@cs.cornell.edu, kqw4@cornell.edu}~, Lequn Wang\footnotemark[1]~~\footnotemark[2]~, Yurong You\footnotemark[1]~~\thanks{Shanghai Jiao Tong University, yurongyou@sjtu.edu.cn}~, Xu Zou\thanks{Tsinghua University, zoux14@mails.tsinghua.edu.cn}~, Vincent  Chen\footnotemark[2]\\
 Serena  Li\footnotemark[2]~, Gao  Huang\footnotemark[2]~, Bharath  Hariharan\footnotemark[2]~, Kilian Q.  Weinberger\footnotemark[2]
$$$$
% For a paper whose authors are all at the same institution,
% omit the following lines up until the closing ``}''.
% Additional authors and addresses can be added with ``\and'',
% just like the second author.
% To save space, use either the email address or home page, not both
}
	
\maketitle
%\thispagestyle{empty}

%%%%%%%%% ABSTRACT
\begin{abstract}
Not all people are equally easy to identify: color statistics might be enough for some cases while others might require careful reasoning about high- and low-level details.
However, prevailing person re-identification(re-ID) methods use one-size-fits-all high-level embeddings from deep convolutional networks for all cases.
This might limit their accuracy on difficult examples or makes them needlessly expensive for the easy ones.
To remedy this, we present a new person re-ID model that combines effective embeddings built on multiple convolutional network layers, trained with deep-supervision.
On traditional re-ID benchmarks, our method improves substantially over the previous state-of-the-art results on all five datasets that we evaluate on.
We then propose two new formulations of the person re-ID problem under resource-constraints, and show how our model can be used to effectively trade off accuracy and computation in the presence of resource constraints. Code and pre-trained models are available at \url{https://github.com/mileyan/DARENet}.
%We also show that it can adapt to resource constraints on three novel formulations of the person re-identification problem.
%mfocus on achieving high accuracies with the assumption of resource availability. In this paper, we make novel use of Deep Supervision to address two new, yet realistic personal re-identification(re-ID) scenarios not addressed previously. The two settings are: 1. \textit{anytime} identification, where prediction for a query can be made available at any stage of the model; and 2. \textit{batch budgeted }re-ID, where limited computation resource is allocated to different queries within a batch depending on difficulty. Specifically, we make use of the triplet loss at multiple stages of a convolutional network that not only learns highly discriminatory embeddings for accuracy but also allows for early-exiting of predictions for efficiency. Experiments on popular re-ID datasets including Market-1501, MARS, and CUHK03 show improvements on the current state-of-the-art in both settings for our proposed framework.

\end{abstract}

%%%%%%%%% BODY TEXT
\vspace{-2ex}
\section{Introduction}
\begin{figure}
\centering
\includegraphics[height=0.5\linewidth]{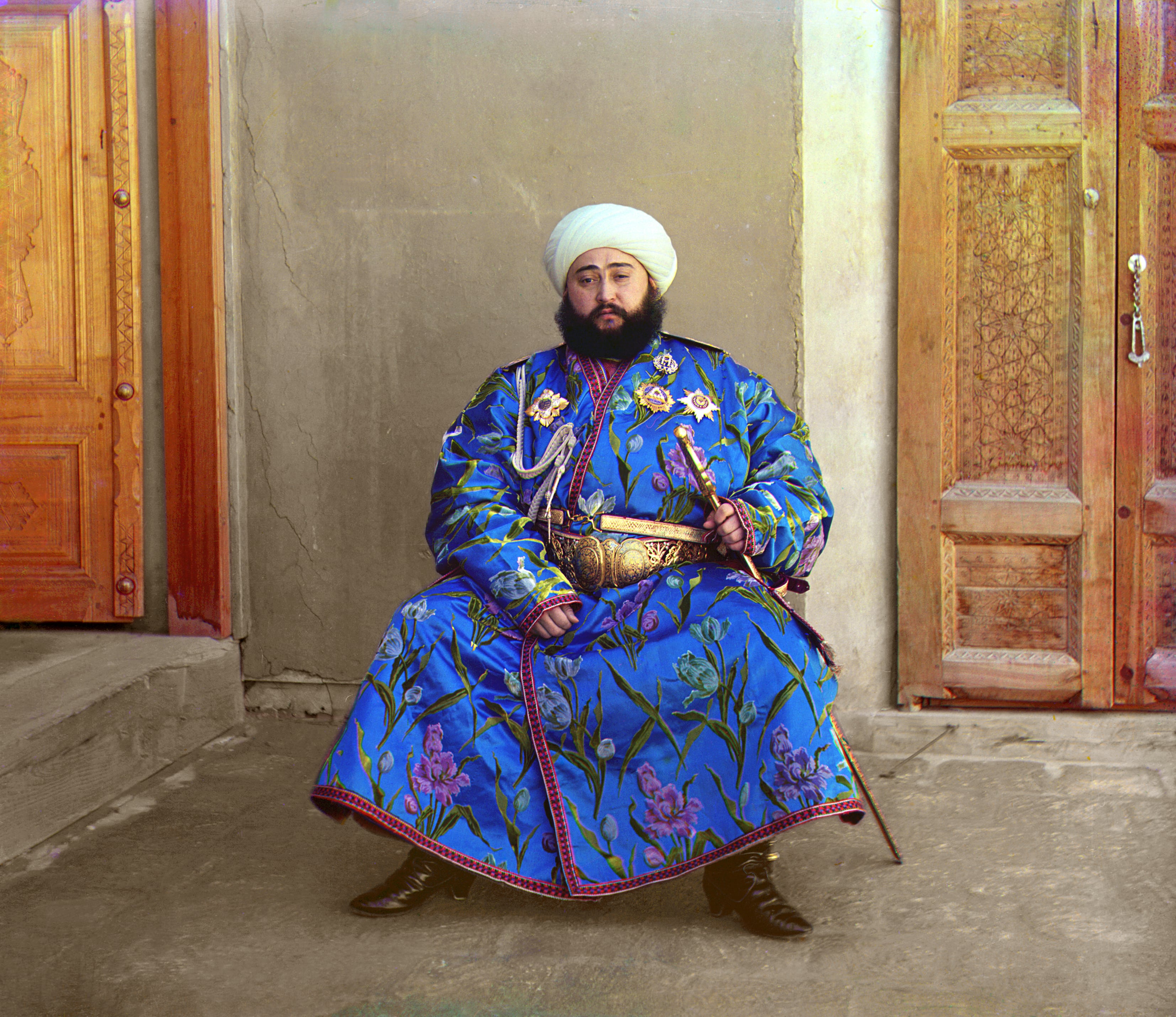}
\includegraphics[height=0.5\linewidth]{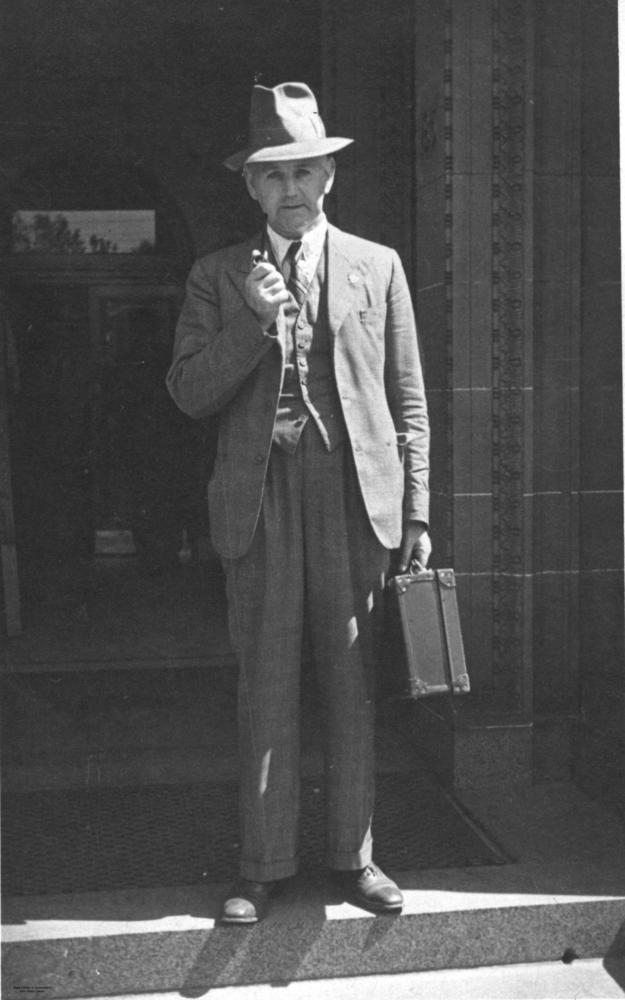}
\caption{Some people have distinctive appearance and are easy to identify (left), while others have nondescript appearance and require sophisticated reasoning to identify correctly (right).}
\vspace{-3ex}
\label{fig:1}
\end{figure}

\vspace{-1ex}
\label{sec:introduction}
Consider the two men shown in Figure~\ref{fig:1}.
The man on the left is easier to identify: even from far away, or on a low-resolution photograph, one can easily recognize the brightly colored attire with medals of various kinds.
By contrast, the man on the right has a nondescript appearance.
One might need to look closely at the set of the eyes, the facial hair, the kind of briefcase he is holding or other such subtle and fine-grained properties to identify him correctly.

Current person re-identification(re-ID) systems treat both persons the same way.
Both images would be run through deep convolutional neural networks (CNNs). Coarse-resolution and semantic embeddings from the last layer would be used to look the image up in the database.
However, this kind of an architecture causes two major problems: first, for the hard cases such as the man on the right in Figure~\ref{fig:1}, these embeddings are too coarse and discard too much information.
Features from the last layer of a CNN mostly encode semantic features, like object presence~\cite{BharathCVPR2015}, but lose all information about the fine spatial details such as the pattern of one's facial hair or the particular shape of one's body.
%Features from the first layers contain high resolution details, but miss the higher order semantics.
Instead, to tackle both cases, ideally we would want to reason \emph{jointly} across multiple levels of semantic abstraction, taking into account both high-resolution (shape and color), as well as highly semantic details (objects or object parts).

In contrast, for the easy cases such as the man on the left in Figure~\ref{fig:1}, using a 50-layer network is overkill.
A color histogram or the low-level statistics computed in the early layers of the network might work just as well. This may not be a problem if all we are interested in is the final accuracy.
However, sometimes we need to be more resource efficient in terms of time, memory, or power.
For example, a robot might need to make decisions within a time limit, or it may have a limited battery supply that precludes the running of a massive CNN on every frame.

Thus standard  CNN-based person re-ID systems are only \emph{one} point on a spectrum.
On one end, early layers of the CNN can be used to identify people quickly under some resource constraints, but might sacrifice accuracy on hard images.
On the other end of the spectrum, highly accurate person re-ID might require reasoning across multiple layers of the CNN.
Ideally, we want a \emph{single} model that encapsulates the entire spectrum. 
This can allow downstream applications to choose the right trade-off between accuracy and computation.

In this paper we present such a person re-ID system.
Our model has a simple architecture, consisting of a standard base network with two straightforward modifications.
First, embeddings across multiple layers are combined into a single embedding.
Second, embeddings at each stage are trained in a supervised manner for the end task.
While both ideas have appeared before in various forms for object detection and segmentation~\cite{xie2015holistically,FarabetTPAMI2013,BharathCVPR2015}, we show for the first time the benefit of these ideas for person re-ID problems, and connect these ideas to the goal of performance under resource constraints.

We evaluate our approach on five well-known person re-ID benchmark datasets. Not only does our method outperform all previous approaches across all datasets, it is also to our knowledge the first person re-ID algorithm applicable to the resource budget settings in test time.

%\yy{should we emphasize the simplicity of our model?}

\vspace{-1ex}
\section{Related Work}
\vspace{-1ex}
\label{sec:RelatedWork}
We briefly review prior work on person re-ID and deep supervision.

\subsection{Person re-ID}
\vspace{-1ex}
%\lw{ some works considered multi-scale features\cite{liu2016multi,li2015multi,fumulti} but in a very different way to ours. Need to tell a better story for that}
Traditional person re-ID methods first extract discriminative hand-crafted features that are robust to illumination and viewpoint changes~\cite{liao2015person,gray2008viewpoint,farenzena2010person,kviatkovsky2013color,zhao2014learning,ma2012local,matsukawa2016hierarchical}, and then use metric learning ~\cite{liao2015person,koestinger2012large,hirzer2012relaxed,dikmen2010pedestrian,zheng2013reidentification,pedagadi2013local, gray2007evaluating, li2013learning,liao2015efficient,lisanti2015person,zhao2013person,bai2017scalable,xiong2014person} to ensure that features from the same person are close to each other while from different people are far away in the embedding space.
Meanwhile, researchers have worked on creating ever more complex person re-ID datasets~\cite{zheng2015scalable,ristani2016performance,li2014deepreid,zheng2016mars} to imitate real-world challenges.

Inspired by the success of CNNs~\cite{lecun1990handwritten} on a variety of vision tasks, recent papers have employed deep learning in person re-ID~\cite{li2014deepreid,ahmed2015improved,wang2016joint,xiao2016learning,cheng2016person,li2017person,zheng2017pedestrian,zheng2017unlabeled,liu_2017_qan,lin2017consistent}. CNN-based models are on the top of the scoreboard. This paper belongs to this large family of CNN-based person re-ID approaches.

There are three types of deep person re-ID models: classification, verification, and distance metric learning.
Classification models consider each identity as a separate class, converting re-ID into a multi-class recognition task~\cite{Zheng2016PersonRP,xiao2016learning,Sun_2017_ICCV}.
Verification models~\cite{li2014deepreid,Yi2014DeepML,Varior2016GatedSC} take a pair of images as input to output a similarity score determining whether they are the same person.
A related class of models learns distance metrics~\cite{ding2015deep,cheng2016person,shi2016embedding,hermans2017defense,chen2017beyond} in the embedding space directly in an expressive way. Hermans et al.~\cite{hermans2017defense} propose a variant of these models that uses the triplet loss with batch hard negative and positive mining to map images into a space where images with the same identity are closer than those of different identities.
We also utilize the triplet loss to train our network, but focus on improvements to the architecture.
Combinations of these loss functions have also been explored~\cite{Chen2017AMD,liu_2017_qan,fumulti}.

Instead of tuning the loss function, other researchers have worked on improving the training procedure, the network architecture, and the pre-processing.
In order to alleviate problems due to occlusion, Zhong et al. \cite{zhong2017random} propose to randomly erase some parts of the input images as the antidote.
Treating re-ID as a retrieval problem, re-ranking approaches \cite{zhong2017re} aim to get robust ranking by lifting up the k-reciprocal nearest neighbors. Under the assumption that correlated weight vectors damp the retrieval performance, Sun et al. \cite{Sun_2017_ICCV} attempt to de-correlate the weights of the last layer. These improvements are orthogonal to our proposed approach. In fact, we integrate random erasing and re-ranking into our approach for better performance.

Some works explicitly consider local features or multi-scale features in the neural networks~\cite{zhao2017deeply,li2017learning,su2017pose,zhao2017spindle,liu2016multi,li2015multi,fumulti}.
By contrast, we implicitly combine features across scale and abstraction by tapping into the different stages of the convolutional network.

\subsection{Deep supervision and skip connections}
\vspace{-1ex}
The idea of using multiple layers of a CNN has been explored before.
Combining features across multiple layers using \emph{skip connections} has proved to be extremely beneficial for segmentation~\cite{FarabetTPAMI2013, BharathCVPR2015, LongCVPR2015} and object detection~\cite{LinCVPR2017}.
In addition, prior work has found that injecting supervision by making predictions at intermediate layers improves performance. This \emph{deep supervision} improves both image classification~\cite{lee2015deeply} and segmentation~\cite{xie2015holistically}. We show that the combination of deep supervision with distance metric learning leads to significant improvements in solving person re-ID problems.

We also present that, under limited resource, accurate prediction is still possible with deep supervision and skip connections. In spite of the key role that efficiency of inference plays in real-world applications, there is very little work incorporating such resource constraints, not even in general image classification setting (exception:~\cite{huang2017multi}).

\vspace{-1ex}
\section{Deep supervision for person re-ID}
\vspace{-1ex}
\label{sec:Method}
\begin{figure*}[!htb]
	\begin{center}
	\includegraphics[width=1.0\linewidth]{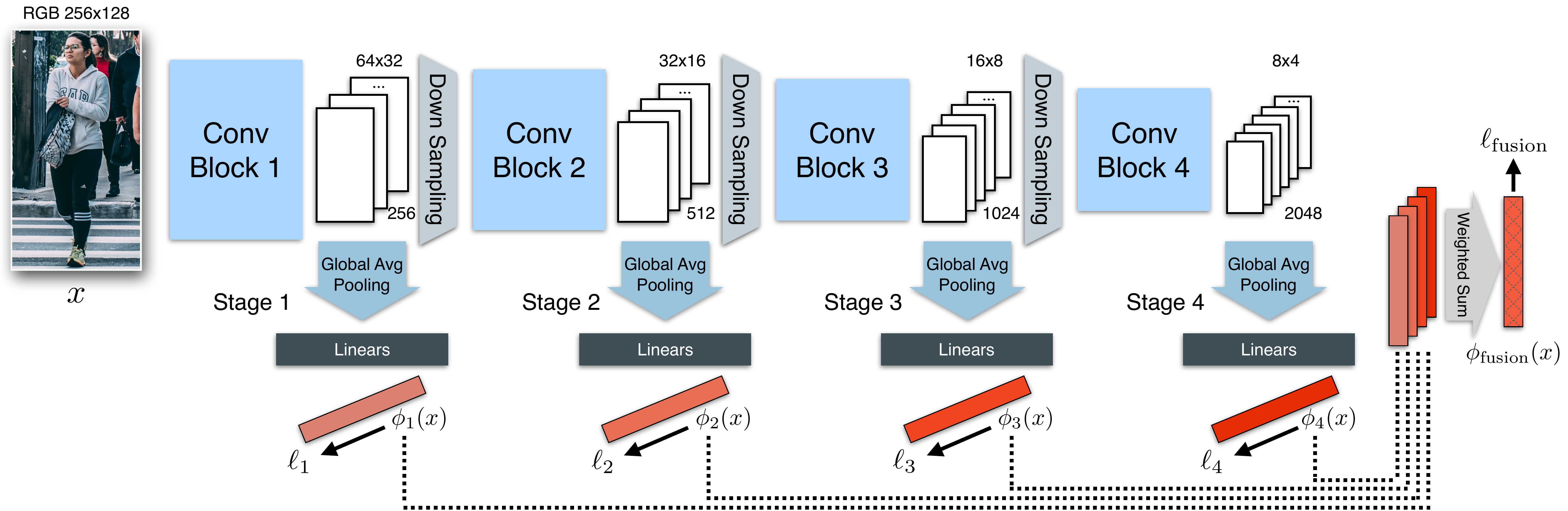}
	\end{center}
\vspace{-1ex}
	    \caption{Illustration of \name{} (\nameshort{}) for person re-ID. The model is based on ResNet-50 \cite{he2016deep} which consists of four \emph{stages}, each with decreasing resolution. \nameshort{} adds extra global average pooling and fully connected layers right after each stage starting from stage 1 (corresponding to conv\_2-5x in \cite{he2016deep}). Different parts are trained jointly with loss $\ell_{all} = \sum_{s=1}^4\ell_{s} + \ell_{\textrm{fusion}}$. When inferring under constrained-resource settings, \nameshort{} will output the most recent available embeddings from intermediate stages (and the ensemble embedding when computation resource is enough for a full pass of the network). (Example image copyright Kaique Rocha (CC0 License)).}
\vspace{-3ex}
	\label{fig:arch}
\end{figure*}

We first consider the traditional person re-ID setting.
Here, the system has a \emph{gallery} $\mathcal{G}$ of images from different people with known identities.
It is then given a \emph{query}/\emph{probe} image $q$ of an unidentified person, which can also be multiple images. The objective of the system is to match the probe with image(s) in the gallery to identify that person.
%In this section, we propose a Deeply Supervised Triplet Network for this problem.
%We first introduce the network architecture which facilitates accurate prediction and then discuss the objective function for training.
%In most re-identification systems, the query $q$ and the gallery images are all projected (e.g., using a neural network) to a common feature space, where the nearest neighbour search can be applied to find the most likely person from the gallery.\lw{Verification Loss does not use nearest neighbor search}

%\subsection{Network architecture}
Previous approaches to person re-ID only use the most high level features to encode an image, e.g., outputs of the last convolution layer form the ResNet-50~\cite{he2016deep}.
Although high-level features are indeed useful in forming abstract concepts for object recognition, they might discard low-level signals like color and texture, which are important clues for person re-ID.
Furthermore, later layers in CNNs are at a coarser resolution, and may not see fine-level details such as patterns on clothes, facial features, subtle pose differences etc.
This suggests that person re-ID will benefit from \emph{fusing} information across multiple layers.

However, such fusion of multiple features will only be useful if each individual feature vector is discriminative enough for the task at hand.
Otherwise, adding in uninformative features might end up adding noise and degrade task performance.

With this intuition in mind, we introduce a novel architecture for person re-ID, which we refer to as \emph{\name{} (\nameshort{})},
 as illustrated in Figure ~\ref{fig:arch}.
Compared to prior work on person re-ID, the architecture a) fuses information from multiple layers~\cite{FarabetTPAMI2013, BharathCVPR2015}, and b) has intermediate losses that train the embeddings from different layers (\emph{deep supervision}~\cite{xie2015holistically}) for person re-ID directly with a variant of the triplet loss.

\vspace{-1ex}
\subsection{Network architecture}
\vspace{-1ex}
Our base network is a residual network (ResNet50)~\cite{he2016deep}.
This network has four \emph{stages}, each halves the resolution of the previous.
Each stage contains multiple convolutional layers operating on feature maps of the same resolution.
At the end of each stage, the feature maps are down-sampled and fed into the next layer.

We take the feature map at the end of \emph{each} stage and use global average pooling followed by two fully connected layers to produce an embedding at each stage. The first fully connected layer has 1204 units including batch normalization and ReLU and the second layer has 128 units.
The function of the fully connected layers is only to bring all embeddings to the same dimension.

Given an image $x$, denote by $\phi_s(x)$ the embedding produced at stage $s$.
We fuse these embeddings using a simple weighted sum:
\vspace{-1ex}
\begin{equation}
\phi_{\textrm{fusion}}(x) = \sum\nolimits_{s=1}^4 w_s\phi_s(x),
\end{equation}
where the weights $w_s$ are learnable parameters.

\vspace{-1ex}
\subsection{Loss function}
\vspace{-1ex}
The loss function we use to train our network is the sum of \emph{per-stage} loss functions $\ell_s$ operating on the embedding $\phi_s(x)$ from every stage $s$ and a loss function on the final fused embedding $\phi_{\textrm{fusion}}(x)$: ${\ell_{\textrm{all}}} = \sum\nolimits_{s=1}^4\ell_{s} + \ell_{\textrm{fusion}}$.
%\begin{equation}
%{\ell^{\textrm{all}}} = \sum\nolimits_{s=1}^4\ell^{s} + \ell^{\textrm{fuse}}.
%\end{equation}

For each loss function, we use the the triplet loss.
The triplet loss is commonly used in metric learning~\cite{weinberger2009distance,Schroff_2015_CVPR} and recently introduced to person re-ID~\cite{cheng2016person,hermans2017defense}.

The reason for using triplet loss is threefold: 1) It minimizes the nearest neighbor loss via expressive embeddings. 2) The triplet loss does not require more parameters as the number of identities in the training set increases. 3) Since it uses simple Euclidean distances, it can leverage well-engineered fast approximate nearest neighbor search (as opposed to the verification models, which construct feature vectors of pairs~\cite{muja2009fast}).

Specifically, we adopt the triplet loss with batch hard mining and soft margin as proposed in~\cite{hermans2017defense}, which reduces uninformative triplets and accelerates training.
Given a batch of images $X$, of $P$ individuals, the triplet loss takes $K$ images per person and their corresponding identities $Y$ in the following form:
\begin{align}
\ell= \sum_{p=1}^{P} \sum_{k=1}^{K}\ln \Big(1 &+ \exp\Big( \overbrace{\max_{a=1, \dots, K} D\left(\phi(x_p^k), \phi(x_p^a)\right)}^{\text{furthest positive}}  \nonumber \\
  &-  \underbrace{\min_{\substack{q=1, \dots, P \\ b=1, \dots, K \\ q\neq p}} D\left(\phi(x_p^k), \phi(x_q^b)\right)}_{\text{nearest negative}}\Big)\Big) ,
  \end{align}
where $\phi(x_p^k)$ is the feature embedding of person $p$ image $k$ and $D(\cdot, \cdot)$ is the $\textrm{L2}$ distance between two embeddings. The loss function encourages the distance to the furthest positive example to be smaller than to the nearest negative example.

\vspace{-1ex}
\section{Resource-constrained person re-ID}
\vspace{-1ex}
\label{sec:resource-constrained}
%\subsection{Person Re-ID under Resource Constraint}
The availability of multiple embeddings from different stages makes our model especially suitable for re-ID applications under resource constraints.
In this section, we consider the person re-ID problem with limited computational resources and illustrate how \nameshort{} can be applied under these scenarios.

\vspace{-1ex}
\subsection{Anytime person re-ID}
\vspace{-1ex}
\label{sec:desc_anytime}

In the anytime prediction setting \cite{grubb2012speedboost, huang2017multi}, the computational budget for a test example is \emph{unknown a priori}, and the re-ID inference process is subject to running out of computation budget at any time.
Although the anytime setting has hardly been studied for person re-ID, it is a common scenario in many settings. For example, imagine a person re-ID app for mobile Android devices that is supposed to perform at a  fixed frame-rate. There exist over $24,093$ distinct Android devices~\cite{huang2017multi} and it is infeasible to ship different versions of an application for each hardware configuration --- instead one may want to ship a single network that can guarantee a given frame rate on all hardware configurations.
%There are over 20,000
%A practical use-case of anytime prediction is in mobile apps on Android devices:
%in 2015, there existed $24,093$ distinct Android devices, each with its own distinct computational limitations.

Here, a traditional re-ID system is all or nothing: it can only return any result if the budget allows for the evaluation of the full model.

Ideally, we would expect the system to have the anytime property, i.e., it is able to produce predictions early-on, but can keep refining the results when the budget allows.
This mechanism can be easily achieved with \nameshort{}: we propagate the input image through the network, and use the most recent intermediate embedding that was computed when the budget ran out to do the identification.

\vspace{-1ex}
\subsection{Budgeted person re-ID}
\vspace{-1ex}
\label{sec:desc_budgeted_batch}

In the budgeted person re-ID problem, the system runs in an online manner, but it is constrained to only use a budget $B$ \emph{in expectation} to compute the answer. The system needs to decide how much computation to spend on each example as it is observing them one by one. Because it only has to adhere to the budget in expectation, it can choose to spend more time on the hard examples as long as it can process easier samples more quickly.
%One application can be using a drone to find the missing child as soon as cheap as accurate as possible, while the battery is everything for the drone.
% \emph{in expectation}.

%The goal for the re-ID system in such a scenario is to go through the query images with certain accuracy and confidence and as cheap as possible for a given computation budget $B$.
We formalize the problem as following: let $S$ be the number of exits (4 in our case), and $C_s > 0$ the amount of computational cost needed to obtain embedding $\phi_s(q)$ at stage $s$ for a single query $q$ ($C_s \leq C_{s + 1}, \forall s = 1,  \dots, S - 1$). At any stage $s$ for a given query, we can decide to ``exit": stop computation and use the $s$-th embedding to identify the query $q$.
Let us denote  the proportion of queries that exit at stage $s$ as $p_s$, where $\sum_{s=1}^S p_s = 1$.
Thus the expected average computation cost for a single query is $\bar{C} =\sum_{s=1}^S p_s C_s$.

\paragraph{Exit thresholds.}
Given the total number of queries $M$ and the total computation budgets $B$, the parameters $\{p_s\}$ can be chosen such that $\bar{C} \leq B / M$, which represents the computation budget for each query. There are various ways to determine $\{p_s\}$.
In practice we define
\begin{align}
	p_s=\frac{1}{Z} a^{s-1},\label{eq:exita}
\end{align}
where $Z$ is the normalization constant and $a\in[0,\inf )$ a fixed constant. Given the costs $C_1,\dots,C_S$, there is a one-to-one mapping between the budget $B$ and $a$. If there were infinitely many stages, eq.~(\ref{eq:exita}) would imply that a fraction of $a$ samples is exited at each stage. In the presence of finitely many exit stages it encourages an even number of early-exits across all stages.
Given $p_s$, we can compute the conditional probability that an input which has traversed all the way to stage $s$ will exit at stage $s$ and not traverse any further as $q_1=p_1$ and $q_s=\frac{p_s}{1-\sum_{i=1}^{s-1}p_i}$.

Once we have solved for $q_s$, we need to decide which queries exit where.
As discussed in the introduction, query images are not equally difficult.
If the system can make full use of this property and route the ``easier'' queries through earlier stages and ``harder'' ones through latter stages, it will yield a better budget-accuracy trade-off.
We solidify this intuition using a simple distance based routing strategy to decide at which stage each query should exit.

\paragraph{Query easiness.}
During testing, at stage $s$, we would like to exit the top $q_s$ percent of ``easiest'' samples. We approximate how ``easy'' a query $q$ is by considering the distance $d_q$ to its nearest neighbor between the query embedding $\phi_s(q)$ and its nearest neighbor in the gallery of the current stage $s$.
A small distance $d_q$ means that we have likely found a match and thus successfully identified the person correctly.
During testing time we keep track of all previous distances $d_{q'}$ for all prior queries $q'$. For a given query $q$ we check if its distance $d_q$ falls into the fraction $q_s$ of smallest nearest neighbor distances, and if it does exit the query at stage $s$.

If labels are available for the gallery at test time, one can perform a better margin based proxy of uncertainty. For a query $q$ one computes the distance $d_q$ to the nearest neighbor, and $d_q'$, the distance to the second nearest neighbor (with a different class membership than the nearest neighbor). The difference $d_q'-d_q$ describes the ``margin of certainty''. If it is large, then the nearest neighbor is sufficiently closer than the second nearest neighbor and there is little uncertainty. If it is small, then the first and second nearest neighbors are close in distance, leaving a fair amount of ambiguity. If labels are available, we use this difference $d_q'-d_q$ as our measure of uncertainty, and remove the top $q_s$ most certain queries at each stage.

\vspace{-1ex}
\section{Experiments}
\vspace{-1ex}
\label{sec:Experiments}
We evaluate our method on multiple large scale person re-ID datasets, and compare with the state-of-the-art.

%Starting with the datasets we evaluate on, we describe the implementation details and further discuss our results.\\
\renewcommand{\arraystretch}{1.1}
\begin{table*}[t]
\centering
\small
\resizebox{\linewidth}{!}{
\begin{tabular}{lcccccccccc}
\hline
\multicolumn{1}{c}{\multirow{3}{*}{Method}} & \multicolumn{10}{c}{Dataset}                                                                                                                                                       \\ \cline{2-11}
\multicolumn{1}{c}{}                  & \multicolumn{2}{c}{Market}  & \multicolumn{2}{c}{MARS}        & \multicolumn{2}{c}{CUHK03(L)} & \multicolumn{2}{c}{CUHK03(D)} & \multicolumn{2}{c}{Duke} \\ \cline{2-11}
\multicolumn{1}{c}{}                  & Rank-1         & mAP            & Rank-1         & mAP            & Rank-1           & mAP              & Rank-1            & mAP              & Rank-1          & mAP             \\ \hline
%LOMO+XQDA~\cite{zhang2016learning}    & 43.8          & 22.2          & -              & -              & 14.8            & 13.6            & 12.8             & 11.5            & 30.7           & 17.0           \\
CNN+DCGAN(R)~\cite{zheng2017unlabeled}   & 56.2          & 78.1          & -              & -              & -                & -                & -                 & -                & 67.7           & 47.1           \\
ST-RNN(C)~\cite{zhou2017see}             & -              & -              & 70.6           & 50.7           & -                & -                & -                 & -                & -               & -               \\
MSCAN(C)~\cite{li2017learning}           & 80.3          & 57.5          & 71.8          & 56.1          & -                & -                & -                 & -                & -               & -               \\
PAN(R)~\cite{zheng2017pedestrian}        & 82.2          & 63.3          & -              & -              & 36.9            & 35.0            & 36.3             & 34.0            & 71.6           & 51.5           \\
SVDNet(R)~\cite{Sun_2017_ICCV}        & 82.3          & 62.1          & -              & -              & 40.9            & 37.8            & 41.5             & 37.2            & 76.7           & 56.8           \\
TriNet(R)~\cite{hermans2017defense}      & 84.9          & 69.1          & 79.8          & 67.7          & -                & -                & -                 & -                & -               & -               \\
TriNet(R)+RE\text{*}~\cite{zhong2017random}      & -              & -              & -              & -              & 64.3            & 59.8            & 61.8             & 57.6            & -               & -               \\
SVDNet(R)+RE~\cite{zhong2017random}   & 87.1          & 71.3          & -              & -              & -                & -                & -                 & -                & 79.3           & 62.4           \\  \hline
\nameshort(R)                                 & 86.4           & 69.3            & 83.0          & 69.7          & 58.1            & 53.7            & 55.1             & 51.3            & 75.2           & 57.4           \\
\nameshort(R)+RE                              & 88.5         &  74.2           & 82.6          & 71.7          & 64.5            & 60.2            & 61.6             & 58.1            &79.1         & 63.0          \\
\nameshort(De)                                 & 86.0                                  & 69.9                                    &  84.2 & 72.1 & 56.4 & 52.2 & 54.3 & 50.1 & 74.5 & 56.3          \\
\nameshort(De)+RE                              & \textbf{89.0}                         &  \textbf{76.0}                          & \textbf{85.5}                          & \textbf{74.0}          & \textbf{66.1} & \textbf{61.6} & \textbf{63.3} & \textbf{59.0} & \textbf{80.2} & \textbf{64.5}          \\ \hline
IDE(C)+ML+RR~\cite{zhong2017re}       & 61.8          & 46.8          & 67.9          & 58.0          & 25.9             & 27.8            & 26.4             & 26.9            & -               & -               \\
IDE(R)+ML+RR~\cite{zhong2017re}       & 77.1          & 63.6          & 73.9          & 68.5          & 38.1             & 40.3            & 34.7             & 37.4            & -               & -               \\
TriNet(R)+RR~\cite{hermans2017defense}   & 86.7          & 81.1          & 81.2          & 77.4          & -                & -                & -                 & -                & -               & -               \\
TriNet(R)+RE+RR\text{*}~\cite{zhong2017random}   & -              & -              & -              & -              & 70.9            & 71.7            & 68.9             & 69.36            & -               & -               \\
SVDNet(R)+RE+RR~\cite{zhong2017random}& 89.1          & 83.9          & -              & -              & -                & -                & -                 & -                & 84.0           & 78.3           \\ \hline
\nameshort{}(R)+RR                                 & 88.3          & 82.0           & 83.0          & 79.3          & 66.0            & 66.7            & 62.8             & 63.6            & 80.4           & 74.5           \\
\nameshort{}(R)+RE+RR                           & 90.8 & 85.9 & 83.9 & 80.6 & 72.9   & 73.7   & 69.8    & 71.2   & \textbf{84.4}  & 79.6  \\
\nameshort{}(De)+RR                                 & 88.6    & 82.2 &  84.8 & 80.3  & 63.4 & 64.1 & 60.2 & 61.6 & 79.7 & 73.3          \\
\nameshort{}(De)+RE+RR                           & \textbf{90.9} & \textbf{86.7} & \textbf{85.1} & \textbf{81.9}  & \textbf{73.8} & \textbf{74.7} & \textbf{70.6} & \textbf{71.6} & \textbf{84.4} & \textbf{80.0}  \\ \hline
%DSNet+SE+RE                               & \textbf{89.61} & \textbf{75.09} & \textbf{84.49} & \textbf{73.31} & \textbf{63.64}   & \textbf{60.18}   & \textbf{62.57}    & \textbf{58.49}   & \textbf{80.57}  & \textbf{64.25}  \\
%DSNet+SE+RE+RR                               & \textbf{91.24} & \textbf{86.78} & \textbf{85.35} & \textbf{82.24} & \textbf{72.71}   & \textbf{73.64}   & \textbf{70.43}    & \textbf{71.30}   & \textbf{84.47}  & \textbf{79.76}  \\ \hline

\end{tabular}
}
\medskip
\caption{Rank-1 and mAP comparison of \nameshort{} with other state-of-the-art methods on the Market-1501 (\emph{Market}), MARS, CUHK03 and DukeMTMC-ReID (\emph{Duke}) datasets. Results that surpass all competing methods are
\textbf{bold}. For convenience of the description, we abbreviate CaffeNet to C, ResNet-50 to R, DenseNet-201 to De, Random erasing to RE and Re-ranking to RR. For CUHK03 dataset, we use the new evaluation protocol shown in~\cite{zhong2017re}, where L stands for hand labeled and D for DPM detected. \text{*} denotes that the result was obtained by our own re-implementation, which yields higher accuracy than the original result.}
%\yy{Remember to add the results from our reimplementation version. Still feel the last column does not look well...} \yy{Should double check the results here.} \sli{1.TriNet should use best result from our implementation 2. Add SD. 3. highlight everything outperform the state of art and mark the best one}}
\vspace{-3ex}
\label{main_results}
\end{table*}

\paragraph{Datasets and evaluation metrics:}
Table~\ref{dataset} describes the datasets used in our experiments.
The images in both Market-1501~\cite{zheng2015scalable} and MARS~\cite{zheng2016mars} are collected by 6 cameras (with overlapping fields of view) in front of a supermarket. Person bounding boxes are obtained from a DPM detector\cite{felzenszwalb2008discriminatively}. Each person is captured by two to six cameras.
%The base elements of Market and MARS are image and traclet respectably.
The images in CUHK03~\cite{li2014deepreid} are also collected by 6 cameras, but without overlapping. The bounding boxes are either manually labeled or automatically generated. The DukeMTMC-reID~\cite{zheng2017unlabeled} contains 36,411 images of 1,812 identities from 8 high-resolution cameras. Among them, 1,404 identities appear in more than two cameras, while 408 identities appear in only one camera.
On all datasets, we use two standard evaluation metrics: rank-1 \emph{Cumulative Matching Characteristic} accuracy (Rank-1) and mean average precision (mAP)~\cite{zheng2015scalable}.
On the CUHK03 dataset, we use the new protocol to split the training and test data as suggested by Zhong et al.~\cite{zhong2017re}. For all datasets, we use the officially provided evaluation code to obtain the results. Our only modification is to use \emph{mean} pooling on the embeddings of a tracklet instead of \emph{max} pooling on MARS.

%The datasets we experimented are standard large-scale deep learning applicable datasets, including the image based Market1501, CUHK03~\cite{li2014deepreid}, DukeMTMC-reID~\cite{zheng2017unlabeled} datasets and the video based MARS~\cite{zheng2016mars} dataset as seen in Table~\ref{dataset}. Here we evaluate our model under standard evaluation metrics including

\renewcommand{\arraystretch}{1.1}
\begin{table}[ht]
\addtolength{\tabcolsep}{0pt}
\centering
\small
\resizebox{\linewidth}{!}{
	\begin{tabular}{l|cccc}
		\hline
		Dataset & Market~\cite{zheng2015scalable} & MARS~\cite{zheng2016mars} & CUHK03~\cite{li2014deepreid} & Duke~\cite{zheng2017unlabeled}\\
		\hline
		Format & Image & Video & Image & Image \\
		Identities & 1,501 & 1,261 &1360 & 1,812\\
		BBoxes & 32,668 & 1,191,003 & 13,164 & 36,411\\
		Cameras & 6 & 6 & 6 & 8\\
		Label method & DPM & DPM+GMMCP  & Hand/DPM & Hand \\
		Train \# imgs & 12,936 & 509,914 & 7,368/7,365 & 16,522\\
		Train \# ids & 751 & 625 &767 & 702\\
		Test \# imgs & 19,732 & 681,089 &1,400& 2,228\\
		Test \# ids & 750 & 635 &700& 702 \\
		\hline
	\end{tabular}
}	
	\medskip
\caption{\label{table1}
The person re-ID datasets used in our experiments. All datasets include realistic challenges, amongst other things due to occlusion, changes in lighting and viewpoint, or mis-localized bounding boxes from object detectors. 
 %This is not necessary: With difficulties like occlusion, multiple-view transform, photometric transform, and the use of automatic detection - deformable part models (DPM)\cite{Forsyth2010ObjectDW} and Generalized Maximum Multi Clique problem (GMMCP)~\cite{dehghanCVPR2015}, those datasets mimic the real-world scenario adequately. CUHK03 is the first person re-ID dataset that is large enough for deep learning, with bounding boxes from hand labeling and DPM. It has better person detection quality than Market-1501 (\emph{Market}). The MARS dataset is an extension version of the Market1501 dataset, which is also the first large-scale video based person re-ID dataset.The DukeMTMC-reID(\emph{Duke}) dataset is a large-scale heavily labeled multi-target multi-camera tracking dataset with 8 cameras.  	
	}
	\label{dataset}
\end{table}

\paragraph{Implementation details:} We use the same settings as in~\cite{hermans2017defense}, except that we train the network for 60,000 iterations instead of 25,000 to ensure a more thorough convergence for our joint loss function (we confirm that training the models in~\cite{hermans2017defense} for more iterations does not help).

Each image is first resized to $256 \times 128$, amplified by a factor 1.125, followed by a $256 \times 128$ crop and a random horizontal flip.
\nameshort{} is built upon a ResNet-50~\cite{he2016deep} or DenseNet-201~\cite{Huang_2017_CVPR} model, pre-trained on ImageNet~\cite{krizhevsky2012imagenet} (both have similar number of parameters). We refer to the two versions as \emph{\nameshort{}(R)} and \emph{\nameshort{}(D)}, respectively. To allow an easier comparison to the TriNet architecture, we performed all experiments in the ablation studies with the ResNet architecture. For notational simplicity we will sometimes drop the $(R)$ in the name and assume that $\nameshort{}$, without specification refers to the ResNet architecture. 
We train both versions of \nameshort{} with Adam~\cite{Kingma2014AdamAM} and a batch size of $72$, which contains $18$ different people, $4$ different images each.
The learning rate $\alpha$  is adjusted similarly as in ~\cite{hermans2017defense}, starting from $\alpha_0\! =\! 3\! \times\! 10^{-4}$
\begin{equation} \label{eq:6}
\alpha(t)=\left\{
\begin{tabular}{ll}
$\alpha_0$ & $\textrm{if} ~ t \leq t_0$,  \\
$\alpha_0 \times 0.001^{\frac{t-t_0}{t_1-t_0}}$ & $\textrm{if} ~ t_0 \leq t \leq t_1$,
\end{tabular}
\right.
\end{equation}
where we set $t_0=30,000$ and $t_1=60,000$. $\beta_1$ in Adam will reduce to 0.5 from 0.9 after $t_0$ as well.
%<<<<<<< HEAD
Following~\cite{hermans2017defense}, the final feature vector of each image during inference is the average over embedding vectors of five crops and their flips~\cite{krizhevsky2012imagenet}. 
All hyperparameters were taken from \cite{hermans2017defense}, optimized for Market and MARS. Potentially we could improve the results of \nameshort{} even further through proper hyperparameter tuning.
% =======

% During inference, the final feature vector of each image is the average over embedding vectors of five crops and their flips~\cite{krizhevsky2012imagenet}.
% This matches the TriNet~\cite{hermans2017defense}.\\
% >>>>>>> 3331f7b8a7b68c6ad930d6cef576f2608710801a

\paragraph{Pre-/post-processing:}
There are two model-agnostic pre- and post-processing steps that increase the accuracy of the person re-ID systems.
\emph{Random erasing} (RE)~\cite{zhong2017random} involves randomly masking parts of the input during training to increase its robustness to occlusion.
\emph{Reranking} (RR)~\cite{zhong2017re} uses the nearest neighbor graph of multiple probe and gallery images to rerank matches.
In our experiments, we evaluate our model both with and without these processing steps.

\subsection{Results on standard person-reID}
%\lw{Should we call this section main results?}

We compare \nameshort{} to existing state-of-the-art methods, and find that \nameshort{} is competitive even without any random erasing or reranking.
The results are shown in Table~\ref{main_results}. We ran several of the experiments four times and found all standard deviations to be less than 0.5.
% on Market, MARS and Duke, \nameshort{} on Market and we show the mean results on the table.
% =======
% The results are shown in Table~\ref{main_results}. We have run four times of \nameshort{}+RE on Market, MARS and Duke, \nameshort{} on Market and we show the mean results on the table. Because the time limitation, for other experiments we only run one time.
% >>>>>>> 3331f7b8a7b68c6ad930d6cef576f2608710801a
In particular, \nameshort{}(R) is uniformly better than TriNet~\cite{hermans2017defense}, which uses the same base network, has a similar number of parameters, and is trained using the same version of triplet loss but without deep supervision or skip connections.
Incorporating random erasing and re-ranking further boosts the results, giving \nameshort{}(R) state-of-the-art performance on all four datasets. 
\nameshort{} is further improved significantly when the base network is changed from ResNet-50 to DenseNet-201.
Our model works well not only on small datasets like Market, CUHK and Duke, but also on large scale datasets like MARS (which contains 1 million images).  Not surprisingly, random erasing tends to improve the  performance substantially on the former, where overfitting can become an issue.

%When we train \nameshort{} on CUHK and Duke datasets, we do not change the settings, which we obtained by~\cite{hermans2017defense} on Market and MARS trainsets, so the performance might increase continually if we do cross-validation on these datasets.
\begin{figure}[t]
	\centering
	\includegraphics[width=\linewidth]{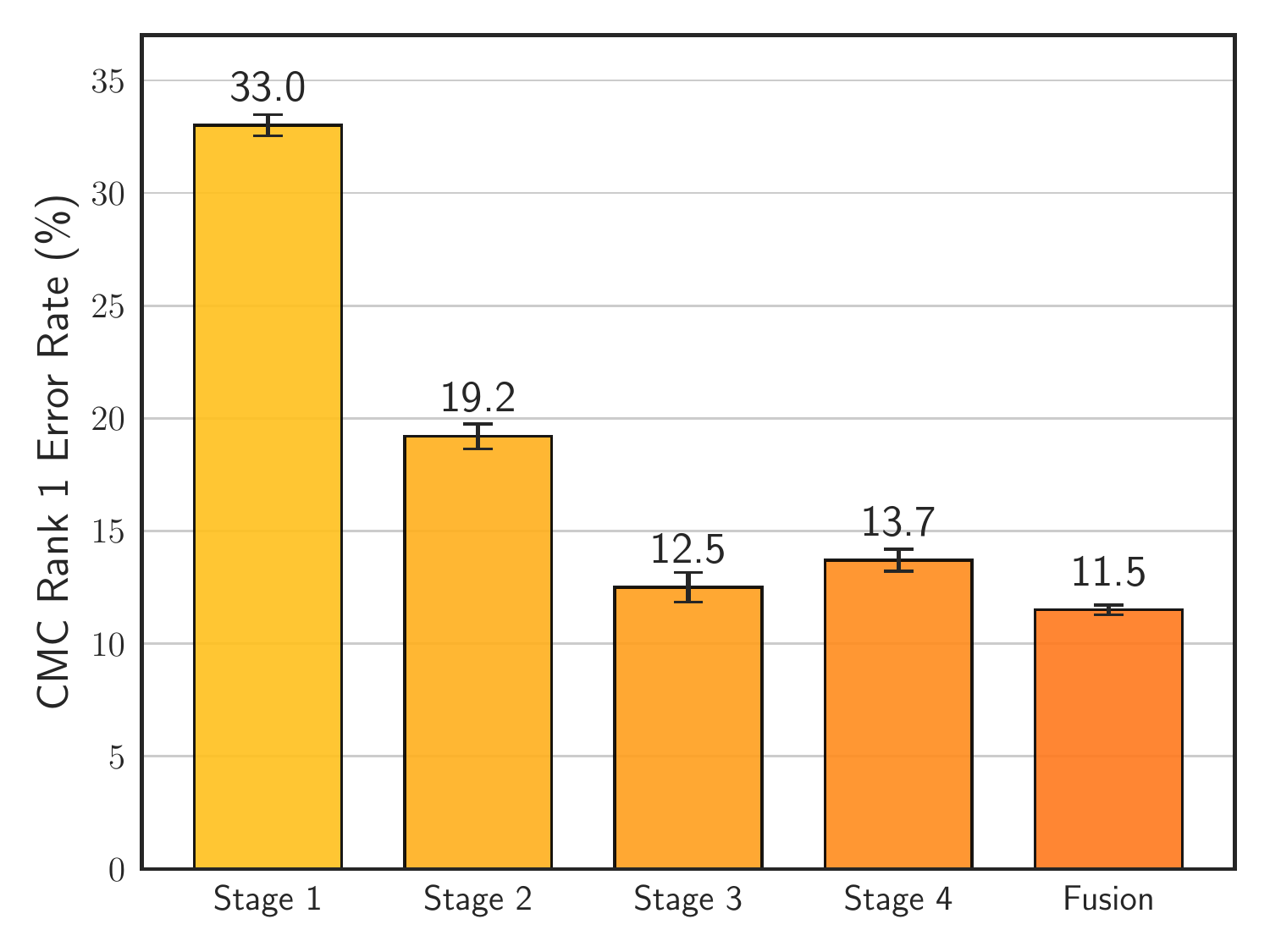}
	\caption{The Rank-1 error on the Market-1501 dataset without re-ranking across the different stages of  \nameshort{} and the final ensemble. Mean and std are estimated over four runs. \nameshort{} is trained with Random Erasing.  }
	\label{different_stages}
\vspace{-2ex}
\end{figure}

\vspace{-1ex}
\subsection{Ablation}
\vspace{-1ex}
The results from Table~\ref{main_results} indicate that \nameshort{} outperforms TriNet significantly.
There are two possible factors behind this improvement: a) the fusion of information from multiple layers, and b) deep supervision.
In the following we analyze the contribution of both factors on the ResNet-50 version of \nameshort{}. 
%Throughout this paragraph, we will denote the ResNet version as \nameshort{} without the (R). 
%In this section, we analyze the cause of such a improvement. There are only two difference between TriNet and DesNet. 1) Fusion 2)Deep supervision.

%bh{This is needless detail}TriNet~\cite{hermans2017defense} is the state-of-the-art method in person re-ID which extracts important information from the image to generate embedding vector. In addition to the same extraction ability, DSNet fuses the extracted features from different stages.The features from the shallow layers contain more detailed information. The features from the deep layers contain more global information. The fusion of these features has more strong representation ability. Further discussed in Section~\ref{sec:bad-case}. The deep supervision part of DSNet includes five lost functions instead of just one comparing to TriNet, so we expect it to be more difficult to train. We use 60,000 iterations to ensure that DSNet converges. To be rigorous about our comparison, we train both TriNet and DSNet with 25,000 and 60,000 iterations, discovering that TriNet is insensitive to more training iterations and DSNet only gain a trivial improvement. The experiment result can be found in Supplement \ref{appendix:A}.
\paragraph{The impact of fusion:}
Figure~\ref{different_stages} shows the performance of the different stages of \nameshort{}, trained with random erasing and evaluated without re-ranking, on the Market-1501 dataset.
As expected, the error rate decreases as one goes deeper into the network and the \emph{fusion} of features from different stages actually achieves the lowest error rate.

However, note that stage 4 achieves lower error rate than stage 3. It is possible that features of the last stage are too ``high level'' and lose too much information due to an extra pooling layer. We further analyze the weights $w_s$ for different stages. When trained with Random Erasing on Market1501, the learnt weights of the four stages are $-0.54,-0.73,-0.77,-0.51$. As expected, the absolute values of the weights of the third stage (the most accurate) is the largest.

Incidentally, note that the early layers of the network also achieve reasonable performance, with even stage 1 reaching a 33.3\% error rate.
This can probably be attributed to deep supervision, which we evaluate next. 
%From the figure, we notice that with deep supervision even stage1 reaches 28.16\% error rate. The deeper features become, the lower error rates achieve, except stage3 to stage4 as the error rate rebound, one explanation can be that the low-resolution features might lose some valuable information. The fusion results achieve the lowest error rate, as combing the different stages information is crucial for person re-ID.

\begin{figure}
\centerline{
\includegraphics[width=0.5\linewidth]{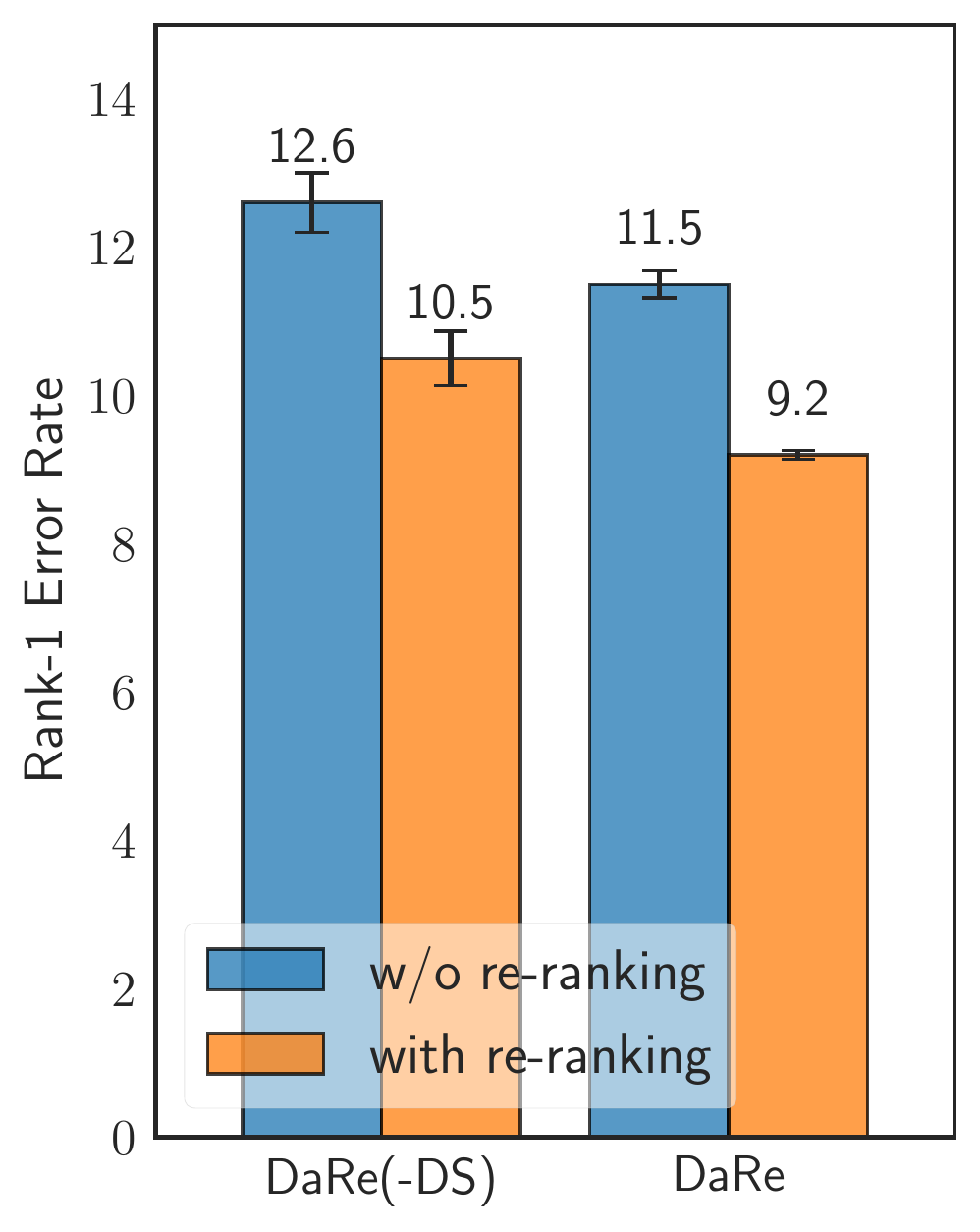}
\includegraphics[width=0.515\linewidth]{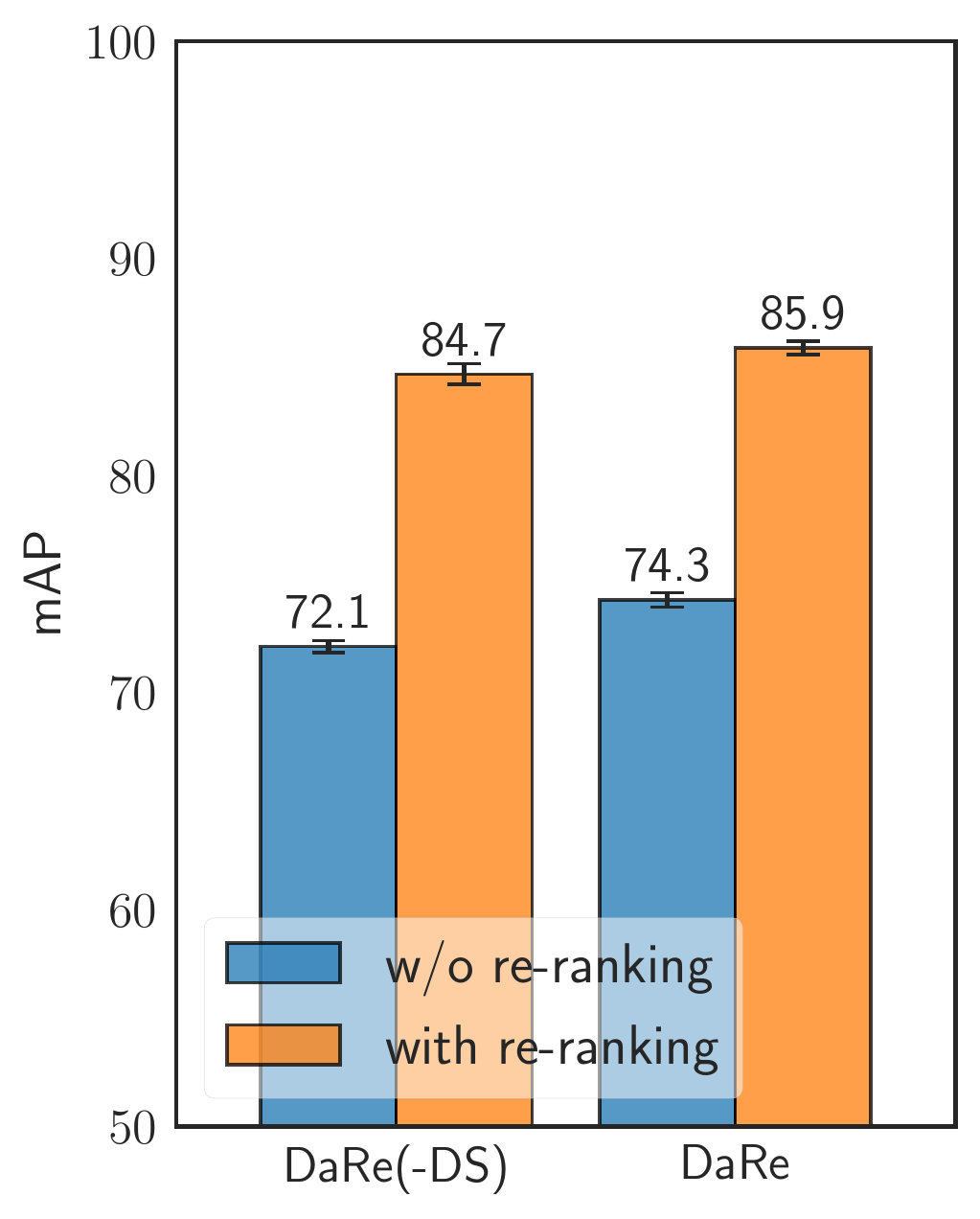}}
\caption{Rank-1 error rates (lower is better) and mAP (higher is better) of \nameshort{} with and without deep supervision on Market1501 dataset. We show means and error bars from four trials.
}
 \vspace{-2 ex}
\label{woensemble}
\end{figure}

\paragraph{The impact of deep supervision:}
We retrain \nameshort{} without any deep supervision, i.e. we remove all intermediate losses except the loss on the fused feature vector.
The results are presented in Figure~\ref{woensemble}.
Without deep supervision, the error rates increase by 2\%, suggesting that deep supervision is indeed required to make sure that each stage learns a good representation.
Intuitively, without deep supervision, the gradients from the loss on the fused feature vector are not informative of how each stage fares individually.
%One conjecture is that deep supervision promotes each stage to learn a better representation. However, if the DSNet has only the fusion, the gradients extended to each stage is decided by nothing but the fusion weight, losing the individual ability to represent.
%Therefore, deep supervision and fusion are both indispensable for \nameshort{}.

\subsection{Results on resource-constrained person re-ID}
\label{sec:exp-anytime-budgeted-batch}

%!TEX root=./egpaper_for_review.tex
We now show results on person re-ID under resource-constrained scenarios described in Section~\ref{sec:resource-constrained}.
All experiments in this section are conducted on Market-1501 dataset.
%Note that this is a new problem and no previous work exists on this problem.
% has attempted to tackle the constrained re-ID.

\vspace{-2ex}
\subsubsection{Anytime person re-ID}
\vspace{-1ex}
%In the anytime person re-ID setting, re-ID system should
\label{sec:exp_anytime}

%\todo{baseline model description for the anytime settings.}
\paragraph{Baselines:} We compare our \nameshort{} model against a sequential ensemble of three ResNets (SE-ResNets)~\cite{he2016deep}, consisting of a ResNet18, a ResNet34 and a ResNet50. 
All three ResNet models are trained separately using the same triplet loss as in \cite{hermans2017defense}.
At test time, the networks are evaluated sequentially in ascending order of size, and are forced to output the most recent re-ID result after surpassing the budget limit.

\begin{figure}[!tb]
	\centering
	\includegraphics[width=\linewidth]{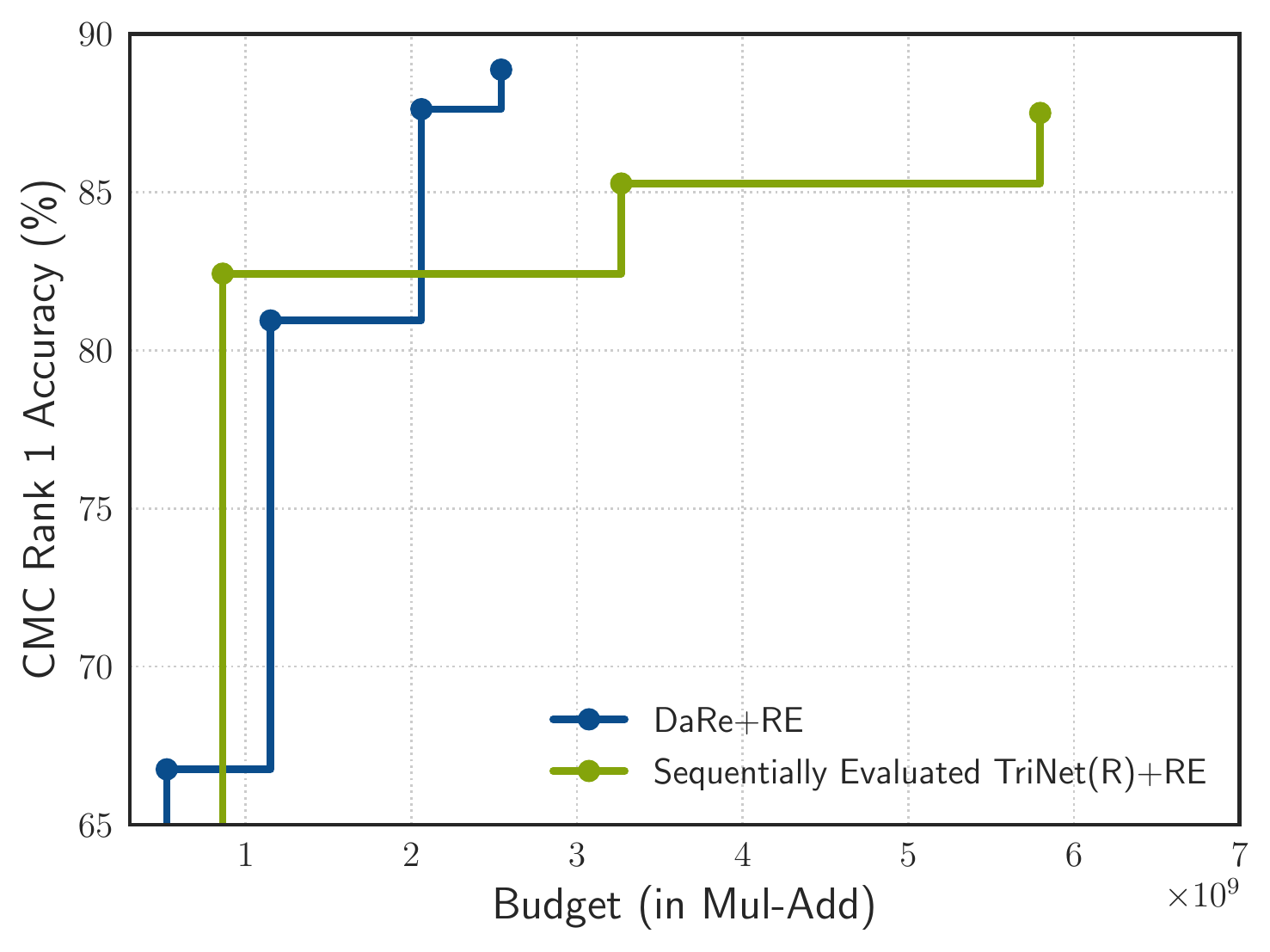}
	\caption{CMC rank 1 accuracy under \textit{anytime re-ID} setting as a function of the  computational budget on the Market-1501 dataset.}
\vspace{-2ex}
	\label{fig:anytime}
\end{figure}

\paragraph{Anytime re-ID results:} Figure~\ref{fig:anytime} summarizes the results of the anytime setting. The computational cost is reported with respect to the cumulative number of  multiplications and additions
(Mul-Add). We confirm that the actual running timing is consistent with the Mul-Add.
Note that we cannot perform re-ranking~\cite{zhong2017re} in this setting since we cannot assume all queries are available at once.
So we report the results from model ``\nameshort{}+RE'' in Table ~\ref{main_results}.

Except for a narrow range of budgets, our \nameshort{} model outperforms the SE-ResNets significantly.
In particular, our model is able to achieve a high accuracy very quickly, achieving $3 \sim 5$ points higher performance for budgets higher than $2.5 \times 10^9$ Mul-Adds.
This is because unlike the SE-ResNets, ours is a single model that \emph{shares} computation between the ``quick-and-dirty'' and the slow-and-accurate predictions.
%Our model is especially able to provide state-of-the-art Especially on budget ranging from $2.5 \times 10^9$ to $5.5 \times 10^9$, our \nameshort{} model achieves $3\sim5\%$ higher CMC rank 1 accuracy.

%\subsubsection{Budgeted batch person re-ID}
%\label{sec:exp_budgeted_batch}
%\paragraph{Confidence - CMC rank 1 accuracy relationship} We show the relationship between the nominal confidence and CMC rank 1 accuracy in Figure \todo{ADD the conf-acc figure}.
%\yy{should I add 1 or all four figure on this topic? since we have four stages.}

%\paragraph{Models}

%\paragraph{Budgeted batch re-ID results} Since no other models have considered such setting, we estimate the computation resource consumption at inference of several state-of-the-art models and plot the corresponding points.
%We also show the performance of the individual embeddings from each stage.
%The distance-based exit strategy outperforms most other models significantly, as seen in Figure ~\ref{fig:budget}.
%Importantly, it is able to correctly trade-off heavy computation on the hard examples with quick prediction on the easy examples, achieving higher accuracy with lower cost than the individual embeddings.
%The LP problem of random exit strategy is solved under $Q^i$ as overall CMC rank 1 at stage $i$, and the confidence-based exit strategy is conducted on choosing $\alpha = \beta = 1$.

%It can be seen that even though we do not incorporate any special design on $\{p_i\}$ under budget $B$, in terms of budget-accuracy trade-off, the distance-based exit strategy outperforms surpassed most other models significantly.

%\begin{figure}[!tb]
%	\centering
%	\includegraphics[width=0.8\linewidth]{./figures/anytime_and_budgeted_batch/budget_batch_reranking}
%	\caption{CMC rank 1 accuracy under \textit{budgeted batch re-ID} setting as a function of computational budget on Market 1501 dataset. Higher is better. Here we follow the same naming in Table.~\ref{main_results}, and use ``\nameshort{}+RE dist-exit'' for the model that adopts the distance-based early exit strategy, ``\nameshort{}+RE'' for model only using one stage output (thus has 4 points in the figure).
%	}
%	\label{fig:budget}
%\end{figure}

%!TEX root=./egpaper_for_review.tex
\vspace{-1ex}
\subsubsection{Budgeted stream person re-ID}
\vspace{-1ex}
\label{sec:stream}

\begin{figure}[!tb]
	\centering
	\includegraphics[width=\linewidth]{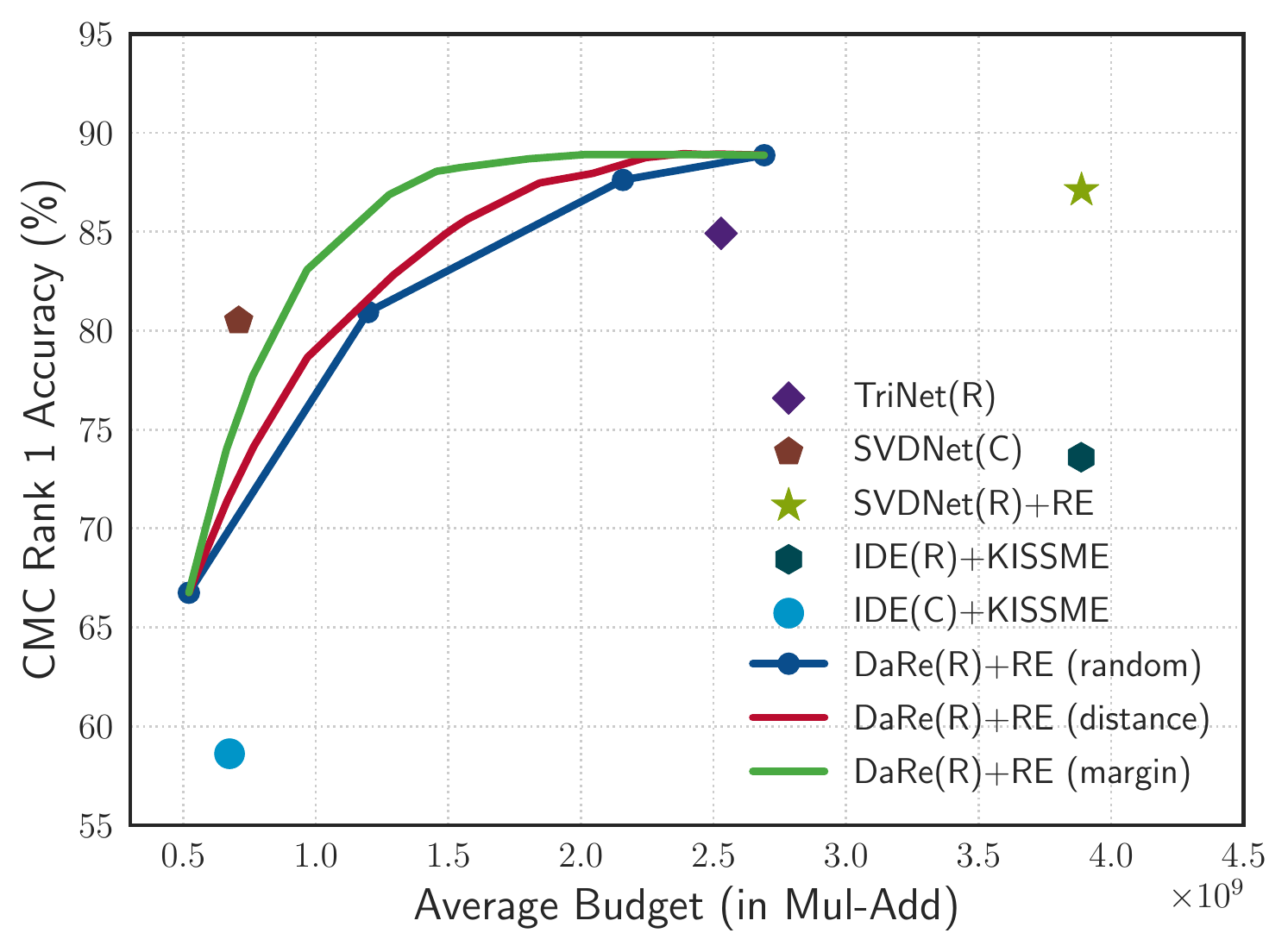}
	\caption{Results on the Market data set under the budgeted streaming setting. The graph shows the CMC Rank 1 accuracy as a function of the average budget in terms of Mul-Add.  }
\vspace{-2ex}
	\label{fig:stream}
\end{figure}
%\paragraph{Budgeted stream re-ID results}
Figure~\ref{fig:stream} shows the results in the budgeted streaming setting. We compare three variants of  four stages \nameshort{}.  Each variant uses a different method to exit queries early. In the \emph{random} setting, we interpolate between the individual stages (indicated as blue points). The straight lines between the blue dots are obtained by randomly deciding to exit queries at either one of the two corresponding stages, which yields a smooth interpolation between the budgets of the two stages. In the \emph{distance} variant, the top $q_s$ queries with the shortest distance to their nearest gallery neighbor are exited at each stage. In the \emph{margin} variant inputs are exited based on their margin of certainty between the nearest neighbor and the second nearest neighbor of a different class. This latter version assumes the knowledge of class labels of gallery data points.
%We compute the Pareto-optimal thresholds on a validation set held out from the training set: $20\%$ of the examples are used for validation.
We observe that our choice of thresholds are able to route queries effectively, allowing us to achieve higher accuracy at lower cost compared to prior state-of-the-art models (which are only single points without the ability to trade-off accuracy for computational cost). When gallery labels are available the \emph{margin} selection method is clearly preferred over the \emph{distance} based method. Both methods  outperform random interpolation.

\begin{figure}[!htb]
	\begin{center}
\vspace{-1ex}
	\includegraphics[width=\linewidth]{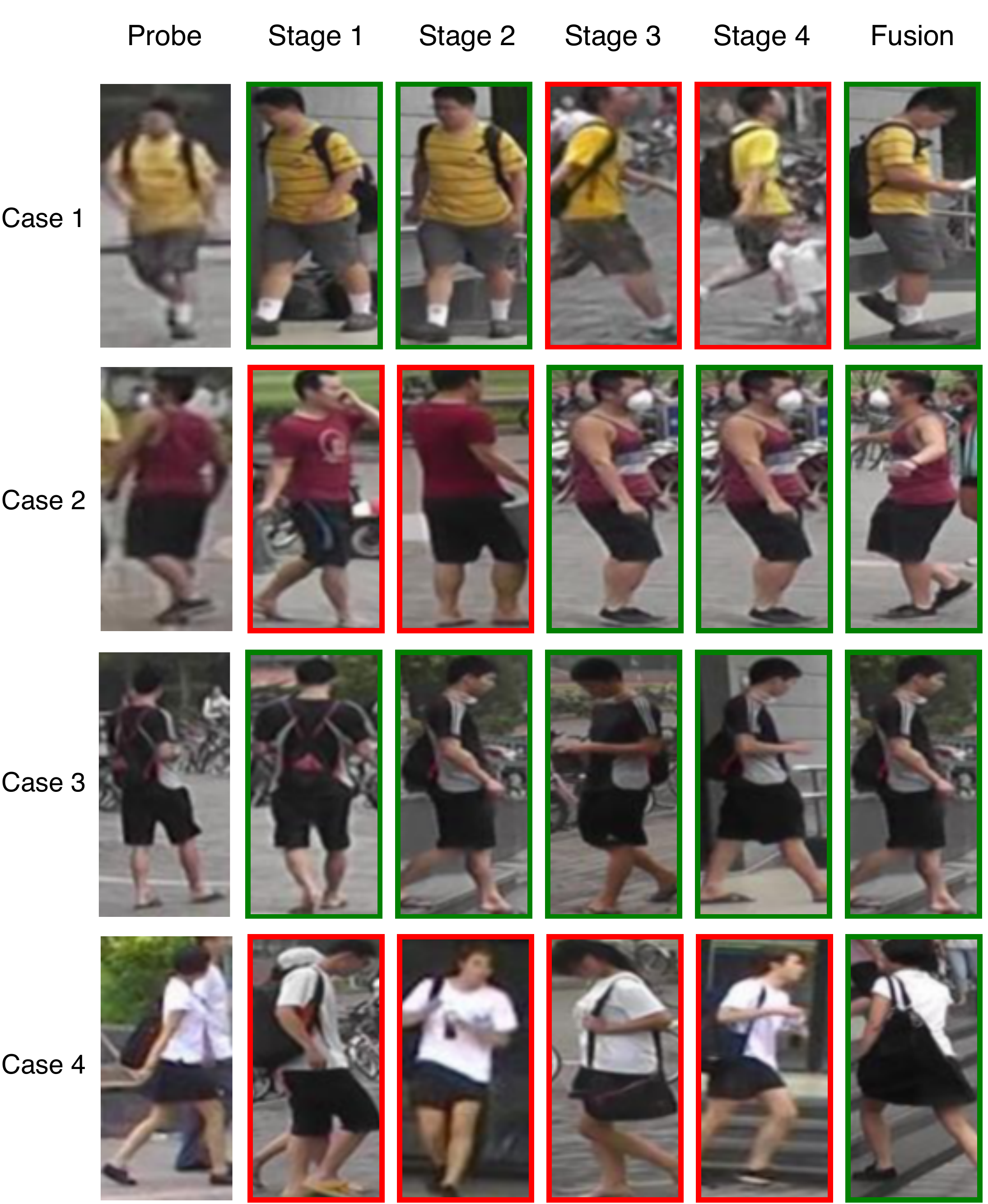}
	\end{center}
	    \caption{Visualization of person re-ID results using features from different stages of our model. We purposely selected samples where the fused feature representation yields the correct results to show where it improves over earlier stages. }
	\label{fig:visualize}
 \vspace{-2 ex}
\end{figure}

\subsection{Qualitative results}
To gain a better understanding of how the features from various stages differ in identifying people, and how the fusion helps, we visualize the retrieved images from four cases in \figurename~\ref{fig:visualize} for which the fused representation classifies the images correctly. The query images are shown in the left most column, and the retrieved images using the features from the four stages of ResNet-50 and the fused embedding are shown in column 2 to 6, respectively. Images with red boxes correspond to wrong identifications, while those with green boxes are correctly identified.

In the four cases, the fused features correctly identify the people from the query image, while low-level (e.g., from State 1) and high-level (e.g., from State 4) features may agree (Case 1, 2) or disagree (Case 3, 4) with each other. In Case 1, the low-level features are more helpful as the stripes on the clothes are important; while in Case 2, they overly emphasize the color signal and produces a wrong identification. In Case 3 and 4, although both low level and high level features yield consistent prediction, they appear to rely on very different information: the former uses more color and texture clues, while the latter seems to use higher level concepts to deal with large variations in pose and view angle. In all cases, the fused feature combines the advantages of both low-level and high-level features and appears to be more reliable than others.

\begin{figure}[!htb]
 	\begin{center}
 	\includegraphics[width=0.9\linewidth]{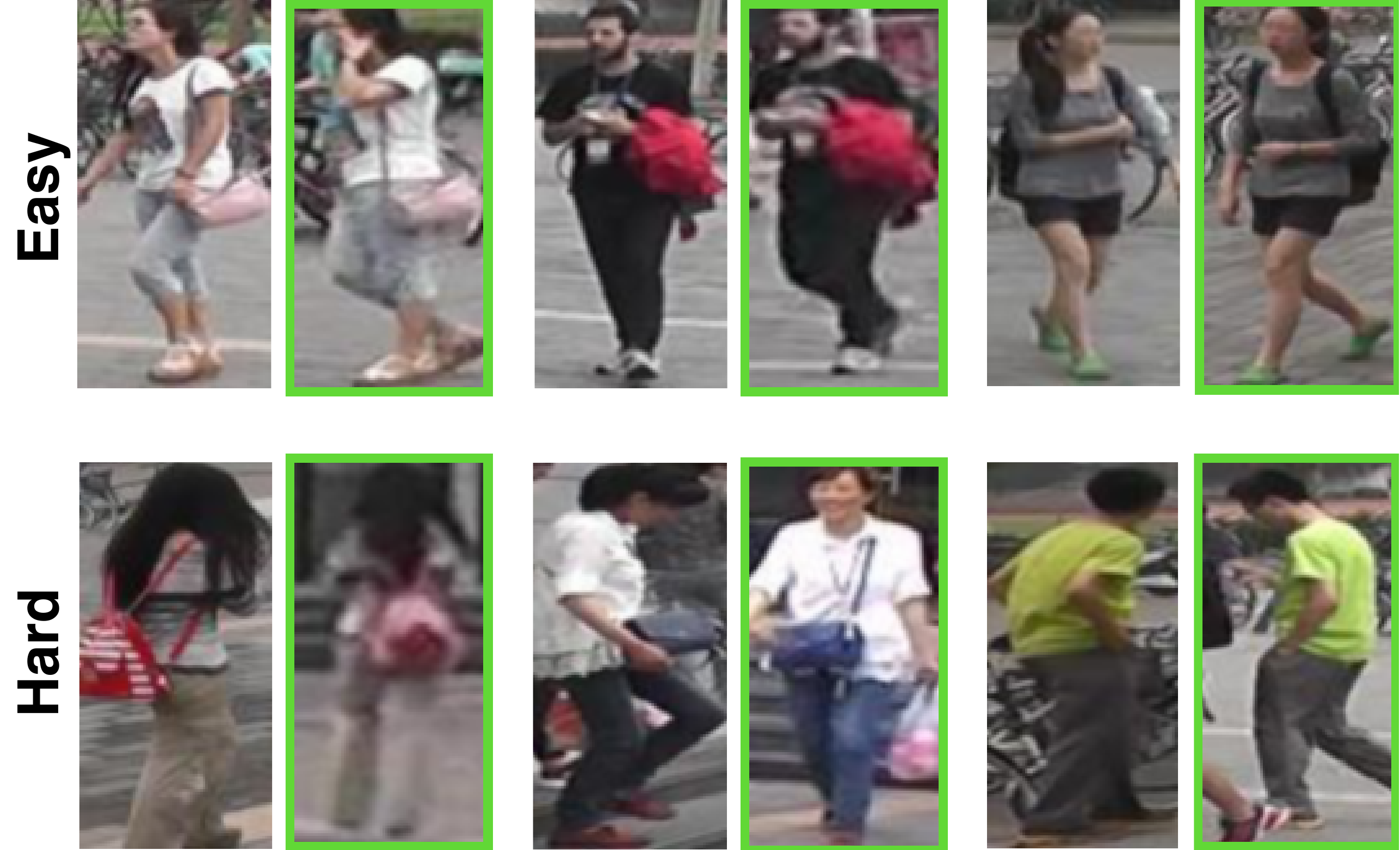}
 	\end{center}
 	    \caption{Visualization of ``easy'' examples, which are confidently classified at the first stage, and ``hard'' examples, which never reach sufficient confidence until the very last stage. }
 	\label{fig:visual-hard}
 \vspace{-2 ex}
 \end{figure}

Figure~\ref{fig:visual-hard}  shows a number of typical query images (together with their matched images from the gallery; green = correct) that are considered to have different difficulties for the network under the budgeted stream re-ID setting. Specifically, the query images (without boxes) in the top row are those exited from the first stage of our model, which we denote as ``easy''. The bottom row shows the ``hard examples'', which are not correctly identified until the last stage of the network. Generally, the separation between easy and hard by the network conforms to our intuitions.

%\section{Discussion}
%\label{sec:Discussion}
%\input{005Discussion}

\section{Conclusion}

\label{sec:Conclusion}
We introduced a novel deeply supervised approach for person re-ID. 
Our model fuses embeddings at both lower (higher resolution) and higher (more semantics) layers of the network. This combination yields achieves state-of-the-art results throughout all our benchmark data sets.
The availability of multiple embeddings with different computation cost also enables trading off performance for computation for the sake of efficiency. As the first work approaching the re-ID problem on a budget efficiency perspective, we show the solutions empirically on the two resource-constrained scenarios using \nameshort{} of  person re-ID.

%It effectively uses both low-level feature patterns, e.g., color and texture, and high-level semantics, e.g., gender and pose, to identify the person in a query image. We showed that this simple approach works surprisingly well in practice, leading to state-of-the-art accuracy on almost all the five competitive datasets we tested on. In addition to being highly accurate, the proposed model can be applied to practical scenarios with computational resource budget. We considered two such settings, namely, the anytime setting and the batch budgeted setting, under both of which our model exhibits remarkable performance in terms of efficiency. We hope the method would become a practically appealing tool for person re-identification, and also provide new insights into the problem.

%------------------------------------------------------------------------

\vspace{-1ex}
\section{Acknowledgements}
\vspace{-1ex}
The authors are supported in part by the National Science Foundation Grants III-1525919,
IIS-1550179, IIS-1618134, S\&AS 1724282, and CCF-1740822, the Office of Naval Research Grant N00014-17-1-2175, and the Bill and Melinda Gates Foundation.

%\clearpage
{\small
\bibliographystyle{ieee}
\bibliography{egbib}
}

%\label{Supplement}
%\input{Supplement}

\end{document}